\documentclass[10pt,twocolumn,letterpaper]{article}

\usepackage{iccv}
\usepackage{times}
\usepackage{epsfig}
\usepackage{graphicx}
\usepackage{amsmath}
\usepackage{amssymb}
\usepackage{algorithm2e, algorithmic}
\usepackage{bm}
\usepackage{mathrsfs}
\usepackage{amsmath}
\usepackage{bm}
\usepackage[cmintegrals]{newtxmath} 
\usepackage{setspace}
\usepackage{authblk}
\usepackage[pagebackref=true,breaklinks=true,letterpaper=true,colorlinks,bookmarks=false]{hyperref}

\iccvfinalcopy % *** Uncomment this line for the final submission

\def\iccvPaperID{1411} % *** Enter the ICCV Paper ID here

\ificcvfinal\fi

\begin{document}

%%%%%%%%% TITLE
\title{Potential Convolution: Embedding Point Clouds into Potential Fields}

\author[1]{Dengsheng Chen}
\author[2]{Haowen Deng}
\author[1]{Jun Li}
\author[3]{Duo Li}
\author[1]{Yao Duan}
\author[1]{Kai Xu\thanks{Corresponding author: kevin.kai.xu@gmail.com}}  

\affil[1]{National University of Defense Technology}  
\affil[2]{Technical University of Munich} 
\affil[3]{HKUST}

%\author{Dengsheng Chen\\
%National University of Defense Technology\\
%Changsha, China\\
%{\tt\small densechen@foxmail.com}
%\and
%Haowen Deng\\
%Technical University of Munich\\
%Munich, Germany\\
%{\tt\small haowen.deng@tum.de}
%\and
%Jun Li\\
%National University of Defense Technology\\
%Changsha, China\\
%{\tt\small jun.johnson.li@gmail.com}
%\and
%Duo Li\\
%HKUST\\
%Hong Kong, China\\
%{\tt\small dlibh@connect.ust.hk}
%\and
%Yao Duan\\
%National University of Defense Technology\\
%Changsha, China\\
%{\tt\small yara.yao.duan@gmail.com}
%\and
%Kai Xu\\
%National University of Defense Technology\\
%Changsha, China\\
%{\tt\small kevin.kai.xu@gmail.com}
%}

\maketitle
\ificcvfinal\thispagestyle{empty}\fi

\begin{abstract}
    Recently, various convolutions based on continuous or discrete kernels for point cloud processing have been widely studied, and achieve impressive performance in many applications, such as shape classification, scene segmentation and so on.
	However, they still suffer from some drawbacks. 
    For continuous kernels, the inaccurate estimation of the kernel weights constitutes a bottleneck for further improving the performance; while for discrete ones, the kernels represented as the points located in the 3D space are lack of rich geometry information. 
    In this work, rather than defining a continuous or discrete kernel, we directly embed convolutional kernels into the learnable potential fields, giving rise to potential convolution.
    It is convenient for us to define various potential functions for potential convolution which can generalize well to a wide range of tasks.
    Specifically, we provide two simple yet effective potential functions via point-wise convolution operations.
    Comprehensive experiments demonstrate the effectiveness of our method, which achieves superior performance on the popular 3D shape classification and scene segmentation benchmarks compared with other state-of-the-art point convolution methods.
\end{abstract}

\section{Introduction}
Convolutional Neural Network~(CNN) plays an important role in boosting a broad range of computer vision applications. Thanks to the regular structure of 2D images, the convolution operations can be done efficiently. However, the world around us is of a higher dimension. With the rapid development of capturing devices, obtaining 3D data is made much easier. As such, the need for processing and understanding 3D data is growing rapidly. As the raw format of 3D data captured by LIDAR sensors, the point cloud is an efficient representation, because it only stores occupied positions in the space. The unordered nature and irregular structure of point clouds make it difficult for researchers to simply apply the traditional convolutions to them. 

Various approaches have been proposed to handle this issue. A straightforward way is to voxelize the space and to represent the 3D data using a grid, then 3D convolutions can be applied accordingly~\cite{maturana2015voxnet,riegler2017octnet,su2018splatnet}. However, the resolution of the volumes is in general limited due to the extra memory and computational overhead and a large number of cells are empty~\cite{graham20183d}. It is thus more desirable to work with raw point clouds directly.

Initial trials were found to work with point-wise convolution, \ie, convolution with the kernel size of 1~\cite{zaheer2017deep, qi2017pointnet}, which inspires a bunch of follow-ups~\cite{wang2018sgpn, landrieu2018large}. Though point-wise convolution can extract point feature with a slight computation burden, it fails to capture local neighborhood information like ordinary convolution operations. 

Efforts on designing special convolution operations targeting point clouds draw much attention in recent years~\cite{atzmon2018point,xu2018spidercnn,li2018pointcnn,hua2018pointwise,hermosilla2018monte,boulch2020convpoint,lin2020fpconv}.
These methods define special convolutions on a set of points in a local neighborhood, sharing the idea that a convolution should contain customizable spatial kernels. 
We can classify current mainstream point convolutions into continuous ones and discrete ones, according to the type of their kernels. 
The continuous ones~\cite{wu2019pointconv} aim to find a mapping function, typically an MLP, which directly convert the local point coordinates to the kernel weights. In other words, the continuous kernels are distributed around the local space and can naturally handle the arbitrary query positions. However, the design of mapping function is always a tricky issue. 
On the contrary, the discrete ones~\cite{thomas2019kpconv} assign an individual position to every kernel and the kernel weights are learned as parameters. By separating positions and weights, discrete point convolution can obtain a more clarified relationship between each input point and kernel. However, it also brings the problem of sparse space coverage of kernels, which makes it sensitive to the sparse point cloud.

Inspired by these findings, we propose a special convolution operation on point clouds, whose kernels are embedded into a set of learnable potential fields which are defined over the whole local space, parameterized by different potential field functions, coined as potential point convolution or potential convolution for short. 
In potential convolution, the kernel weights are learned separately, just like discrete one. The potential convolution kernels are not the discrete points, but instead represented by the learned potential fields around the local space. This makes potential convolution handle arbitrary input point distribution much better. 
Compared with continuous point convolution, potential convolution can directly learn a more accurate weight vector for each potential function without the demand to regress the kernel weights from an MLP, which also relieves the computation burden. 
Compared with discrete point convolution, potential kernels are defined around the whole local space rather than only sited some discrete positions like discrete kernels. Thus, potential convolution is more robust to abnormal distribution.

Depending on the basic conception of potential convolution, we derive the linear and quadratic potential field functions. We also give a further analysis from the geometric aspect and find that the linear one makes an analogy to planes that contain more direction information and the quadratic one makes an analogy to surfaces that contain more complicated shape information.

The naive implementation of potential convolution can be not so memory-efficient and difficult to train. Hence, we propose to simplify it using several point-wise convolution operations with fewer computation cost, which also eases the subsequent training stage for a better performance. 

In a nutshell, the main contributions of this work include:
\begin{itemize}
    \item We come up with a novel design of the point convolution operation based on the potential field, termed as \textit{potential convolution}.
    % We propose a new kind of conception to design point convolution kernels from potential fields.
    \item Two simple yet effective potential functions are proposed to serve as the core of potential convolutions, referred as linear and quadratic respectively. A geometric insight is also provided to explain why they would work. 
    % We propose the linear and quadratic potential function for our potential kernels and achieve state-of-the-art performance on shape classification and scene segmentation tasks.
    \item Our methods obtain state-of-the-art results on a series of point cloud-based vision tasks, including shape classification, shape segmentation and scene segmentation, strongly proving the effectiveness of our potential convolution. 
    % We further demonstrate the relationship between our proposed potential filed and the basic geometric shape and illustrate how our method works.
\end{itemize}

\section{Related work}
In this section, we review deep learning methods in analyzing 3D data, paying particular attention to the methods working on point clouds. Based on the format of the input data, they can be roughly categorized into three groups. 

\paragraph{Image-based}
3D data is projected onto planes to generate a set of 2D images, which can be further processed directly by 2D CNNs~\cite{boulch2017unstructured,lawin2017deep, tatarchenko2018tangent}. For point clouds, however, missing surfaces and density variations have negative impacts on the resulting projections. 

\paragraph{Voxel-based} 
3D data is positioned in a volumetric grid and partitioned by the cells. Each cell records the status of occupancy, occlusion, distance to the surface, \etc~\cite{maturana2015voxnet,roynard2018classification,ben20183dmfv}. This way, the data can be processed directly by CNNs using 3D kernels. However, the extra dimension means a cubic increase of computation and storage as the resolution of the grid increasing. So fine-grained details are often lost after vocalization using large cell size. Some efforts are made to take advantage of the sparsity of the volumes, such as octrees or hash maps~\cite{riegler2017octnet,graham20183d}. 
The idea of sparse convolution also improves the computation efficiency by a large extent~\cite{graham20183d,graham2017submanifold}. Similarly, ~\cite{li2016fpnn} scans the 3D space to decide where computations are necessary.
But their kernels are in general constrained to stay small subject to the large computation burden brought by 3D convolutions.

\paragraph{Point-based}
Raw point clouds can be taken as input directly without converting them to another regular structured format. It is drawing more and more attention as it enjoys full sparsity~\cite{ravanbakhsh2016deep,qi2017pointnet,qi2017pointnet++,su2018splatnet,tatarchenko2018tangent,hua2018pointwise,groh2018flex,verma2018feastnet}. PointNet~\cite{qi2017pointnet,ravanbakhsh2016deep} proposes to use shared multi-layer perceptrons and max pooling layers to extract features of point clouds. PointNet++~\cite{qi2017pointnet++} is an enhanced version of PointNet~\cite{qi2017pointnet} by adding a hierarchical structure. It is similar to the one used in image-based CNNs by extracting features, starting from small local regions and gradually extending to larger regions. The key components used in both PointNet~\cite{qi2017pointnet} and PointNet++~\cite{qi2017pointnet++} to aggregate features from different points are max-pooling layers. However, they keep only the strongest activation of features across a set of points, which may not be the optimal and some useful information might be lost.
Regarding this, researchers strive to find a better way of transforming and aggregating features for point clouds. Different point convolutions are proposed for this purpose. 

Based on the type of kernels used, we further classify them into two categories.
\begin{itemize}
    \item \textbf{Continuous kernel} 
    Unlike discrete kernel, continuous kernel parameterizes a continuous weight space in the local coordinate system. When applied to point clouds, each point will get a specific weight based on its position in the local neighborhood. Yet, to accurately estimate the weight for each point is more challenging and requires more computation~\cite{wu2019pointconv,boulch2020convpoint}.

    \item \textbf{Discrete kernel} 
    The kernel can be viewed as a set of discrete points distributed in the space and each carries a weight. In most cases, the shape of the kernel, \ie, positions of the kernel points, is manually designed and rigid. When applied to point clouds, each point is associated with the closest kernel weight~\cite{hua2018pointwise,xu2018spidercnn,atzmon2018point}. A deformable version which learns to add a shift to each kernel point is found to be more flexible and improve the representation capability~\cite{dai2017deformable,thomas2019kpconv}. However, the performance of such methods will drop drastically if the input points are extremely sparse or abnormal.
\end{itemize}

As mentioned above, both of them have some limitations for point cloud processing. 
The discrete kernels suffer from the discrete and sparse kernel coverage in the entire space, and the continuous kernels suffer from the inaccurate estimation of the weight for each position. 

\begin{figure*}
    \center
    \includegraphics[width=0.9\textwidth]{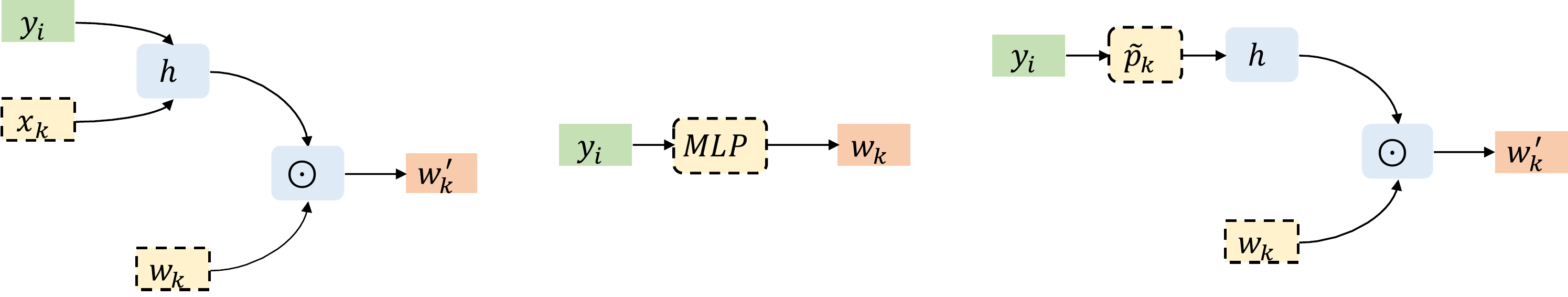}
    \caption{
    Three different kinds of kernel function $g$. 
    Left: the discrete kernel function. The discrete kernel consists of a position vector $\mathbf{x}_k$ and weight vector $\mathbf{w}_k$. The correlation function $h$ takes in $\mathbf{y}_i$ along with $\mathbf{x}_k$ to adjust $\mathbf{w}_k$ and output $w_k^\prime$ for each point $\mathbf{y}_{i}$. The position vector $x_k$ is learnable or predefined at some times.
    Middle: the continuous kernel function. The continuous kernel function is usually represented as a multi-layer perceptron function. It directly takes in $\mathbf{y}_i$ to regress the weight vector $\mathbf{w}_k$ for convolution operator. 
    Right: our proposed potential kernel function. In potential convolution, we are no longer to learn the discrete position $\mathbf{x}_k$ like discrete one, but we wrapper $\mathbf{y}_i$ with a potential field formulated as $\tilde{p}_k$, then apply $h$ on the potential value calculated by $\tilde{p}_k$. 
    The dotted boxes indicate that contains learnable parameters.
    }
    \label{fig:conv}
\end{figure*}

In this work, we drop the conventional notation of discrete and continuous kernel type, and embed kernel into learnable potential fields instead. As shown in Fig.~\ref{fig:conv}, it is not necessary for potential convolution to learn position of each kernel, while only the potential function is needed to be learned. Compared with the kernel points distributed discretely in the local space, our potential field can cover the whole local space and make it more suitable to fit different input points. In the meantime, the computation of potential function is efficient, and we can provide an accurate potential value for each input point. The potential value will act as a factor to adjust the weight vector. The adjusted weight vector contains both geometry and feature information about local points, and is finally applied to the input features.

Once the appropriate potential functions defined, potential convolution can be well adapted into various network structures without any data-dependent hyper-parameters.
We provide two simple but effective potential functions to demonstrate the superiority of our potential convolution, compared with the state of the art continuous/discrete point convolution methods. 
We also give a novel illustration from the geometric aspect to show that why potential convolution works better than discrete one and find that potential kernel contains more complex geometric information. 

\section{Potential Convolution}
\begin{figure}
    \center
    \includegraphics[width=0.45\textwidth]{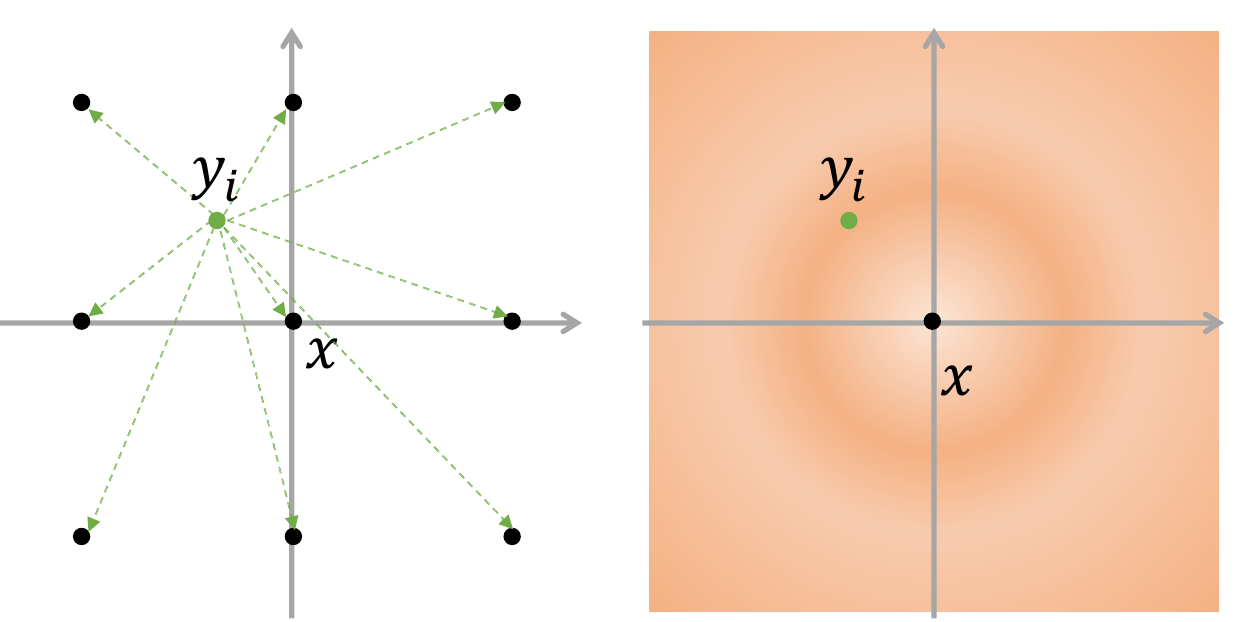}
    \caption{Discrete (left) and continuous (right) kernel type. 
    We project discrete and continuous kernels into two-dimensions for a better visualization.
    Discrete kernels are some points that lies in local space, \ie black points in left plot.
    Continuous kernels can cover the whole local space, \ie the colorized background in right plot.}
    \label{fig:kernel}
\end{figure}

In this section, we present our potential convolution, whose kernel functions are built on top of the potential fields. 
The potential fields have been frequently adopted in robot path planning~\cite{bounini2017modified,park2001obstacle,vadakkepat2000evolutionary}. Each position in the field is granted a potential value via some parameterized function, indicating properties of interest. As such, we import potential fields into our design of kernel functions to quantitatively measure the relationships between points and the learned kernel weights, which are further utilized to aggregate the local information embedded.    

\subsection{A Kernel Function Defined by Potential Fields}

A general definition of a point convolution can be written as
\begin{equation}
    \mathcal{F} \ast g (\mathbf{x}) = \sum_{\mathbf{x}_i \in \mathcal{N}_\mathbf{x}} g(\mathbf{x}_i) \mathbf{f}_i,
\label{eq:pointconv}
\end{equation}
where $\mathcal{N}_\mathbf{x} \in \mathbb{R}^{K\times 3}$ is the set of points in the neighborhood of $\mathbf{x}$ and $\mathcal{F} \in \mathbb{R}^{K\times D}$ stands for the collection of their corresponding features. $K$ is the number of neighboring points and $g$ refers to a specialized kernel function. $\mathbf{x}_i$ and $\mathbf{f}_i$ are members from $\mathcal{N}_\mathbf{x}$ and $\mathcal{F}$ respectively. 

There are different ways of defining a local neighborhood. We opt for the radius-based one to ensure the robustness to varying densities as in~\cite{hermosilla2018monte}. It is also shown by ~\cite{thomas2018semantic} that a better handcrafted 3D features can be obtained this way as opposed to the KNN-based version. Nevertheless, the choice of constructing a neighborhood has only a minor influence on our method as potential convolution pays more attention to the underlying shape information represented by the set of points instead of individual points themselves.

The most important part in Eq.~\ref{eq:pointconv} is the implementation of the kernel function $g$, that is how point convolutions differ from each other. 
Usually, $g$ takes the coordinates of the points in $\mathcal{N}_{\mathbf{x}}$ re-centralized by $\mathbf{x}$ as input, \ie $\mathbf{y}_i=\mathbf{x}_i-\mathbf{x}$. 
% Usually, $g$ takes the neighbors positions centered on $\mathbf{x}$ as input. We call them $\mathbf{y}_i=\mathbf{x}_i-\mathbf{x}$ in the following. 
As our neighborhood is defined by the radius $r$, the input domain of $g$ is the ball $\mathcal{B}_{r}^3=\{\mathbf{y}\in \mathbb{R}^3 | \Vert \mathbf{y} \Vert \le r\}$. 
In the discrete and continuous kernels, $g$ will apply different weights to different areas inside this domain based on their position relationship, as shown in Fig.~\ref{fig:kernel}.
In potential kernels, $g$ will apply different weights to different areas inside this domain based on their potential values, as shown in Fig.~\ref{fig:pot}.
There are many ways to define areas in 3D space used for discrete and continuous convolution, and points are the most intuitive as features are localized by them.
But in potential convolution, we can directly define the potential fields as the whole local 3D space $\mathcal{B}_{r}^3$, and any point $\mathbf{y}_i$ possess its own potential value as the adaptive factor for kernel weights. 

Let $\{ \tilde{p}_{k} | k < D^\prime \} \subset \mathbb{B}_r^3$ be the potential fields lies in local 3D space, $D^\prime$ is the number of potential fields, \ie the output dimension of potential convolution.
Let $\{w_k | k < D^\prime \} \subset \mathbb{R}^{D}$ be the associated weight matrix that map features form dimension $D$ to $D^\prime$. We define the kernel function $g$ for any point $\mathbf{y}_i \in \mathbb{B}_{r}^3$ in each potential field $\tilde{p}_{k}$ as:
\begin{equation}
    g_k(\mathbf{y}_i) =(h \circ \tilde{p}_{k})(\mathbf{y}_i)w_k,
\end{equation}
where $h(\cdot)$ is the correlation function that should be higher when $\mathbf{y}_i$ obtain a lower potential value under $\tilde{p}_k$. In other words, if $\mathbf{y}_i$ gets closer to the zero-potential surface, we think it can better fit into current potential fields, and a higher correlation value should be returned by $h(\cdot)$.

Inspired by~\cite{atzmon2018point}, we also use the Gaussian function to calculate the correlation value:
\begin{equation}
    h(p) = \frac{1}{\sqrt{2\pi}\sigma} exp(- \frac{(p-\mu)^2 }{2\sigma^2}),
\label{eq:h}
\end{equation}
where $\sigma, \mu$ is the variance and mean factor of Gaussian function, and $p=\tilde{p}_{k}(\mathbf{y}_i)$. Rather than ~\cite{thomas2019kpconv}, which used a simpler linear correlation function, we find that Gaussian function benefit learning a more stable potential field $\tilde{p}$. On the other way, we also make a simplification of the standard Gaussian function. We choose $\sigma=\frac{1}{\sqrt{2}}, \mu=0$ and ignore the coefficient terms $\frac{1}{\sqrt{2\pi}\sigma}$, which yields:
\begin{equation}
    h(p) = exp(-p^2).
\label{eq:simh}
\end{equation}
The simplification eases the computation as well as stabilizes the training process. 

\subsection{Linear or Quadratic Potential Field}
Potential functions $\tilde{p}_k$ are critical to the convolution operator. The computation of potential functions should be efficient. 
There are many ways to define different potential functions, even using a small neural network. 
However, it is more convincing for us to adapt a simple potential function than a complex one to demonstrate the powerful feature extraction capabilities of our method.
Hence, we provide two kinds of the simplest potential functions, \ie linear one $\tilde{p}^1_k$ and quadratic one $\tilde{p}^2_k$.

\paragraph{Linear Potential Function}
The linear potential function is defined as:
\begin{equation}
    \tilde{p}_k^1(\mathbf{y}_i) = \mathbf{a}_k\mathbf{y}_i^T+d_k,
\label{eq:pk1}
\end{equation}
where $\mathbf{a}_k \in \mathbb{R}^3$ is learnable parameters of potential function and $d_k \in \mathbb{R}$ is bias. 
Furthermore, we can easily extend this function to a more general version. 
For example, if normal vector $\mathbf{n}_{\mathbf{y}_i}$ is given, we can extend $\tilde{p}_k^1$ to:
\begin{equation}
    \tilde{p}_k^\prime(\mathbf{y}_i, \mathbf{n}_{\mathbf{y}_i})=\mathbf{a}_k(\mathbf{y}_i + \mathbf{n}_{\mathbf{y}_i})^T+d_k.
\label{eq:pk'}
\end{equation}

\paragraph{Quadratic Potential Function}
The quadratic potential function is defined as:
\begin{equation}
    \tilde{p}_k^2(\mathbf{y}_i) = \mathbf{a}_k\mathbf{y}_i^T+\mathbf{b}_k(\mathbf{y}_i^2)^T+d_k,
\label{eq:pk2}
\end{equation}
where $\mathbf{a}_k, \mathbf{b}_k \in \mathbb{R}^3$ is linear and quadratic learnable parameters respectively, $d_k \in \mathbb{R}$ is bias.

\subsection{An Analysis from Geometric Aspect}
\begin{figure}
    \centering
    \includegraphics[width=0.45\textwidth]{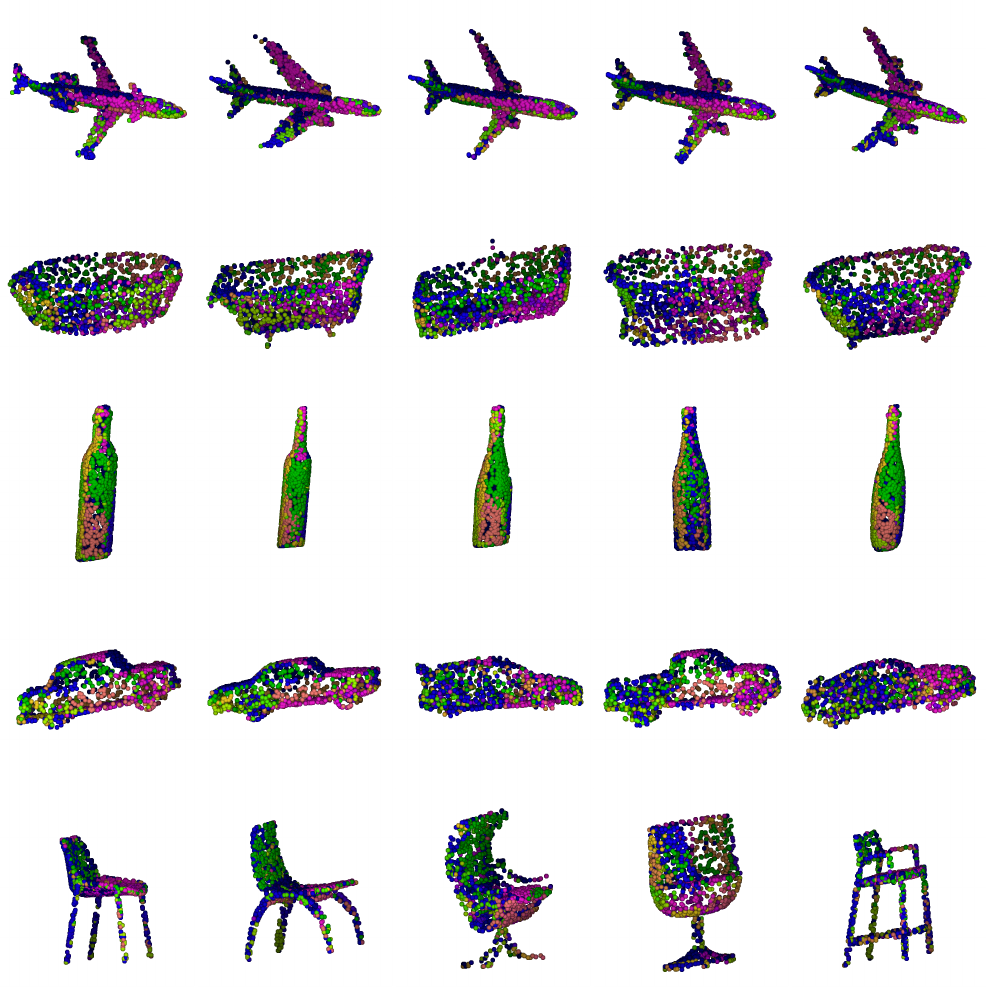}
    \caption{Visualization of learned linear potential fields. Each potential field $\tilde{p}_k$ assigns with a specified color. We colorize the points $\mathbf{y}_i$ with the color of the most related potential fields $\tilde{p}_k$, \ie $\tilde{p}_k={argmax}_{k\in D^\prime}(h \circ \tilde{p}_k)(\mathbf{y}_i)$. As shown in the plot, the points in the same direction are more likely to assign same color, which indicates that the learned linear potential fields can better utilize the direction information of input data.} 
    \label{fig:plane}
\end{figure}

\begin{figure*}
    \centering
    \includegraphics[width=0.8\textwidth]{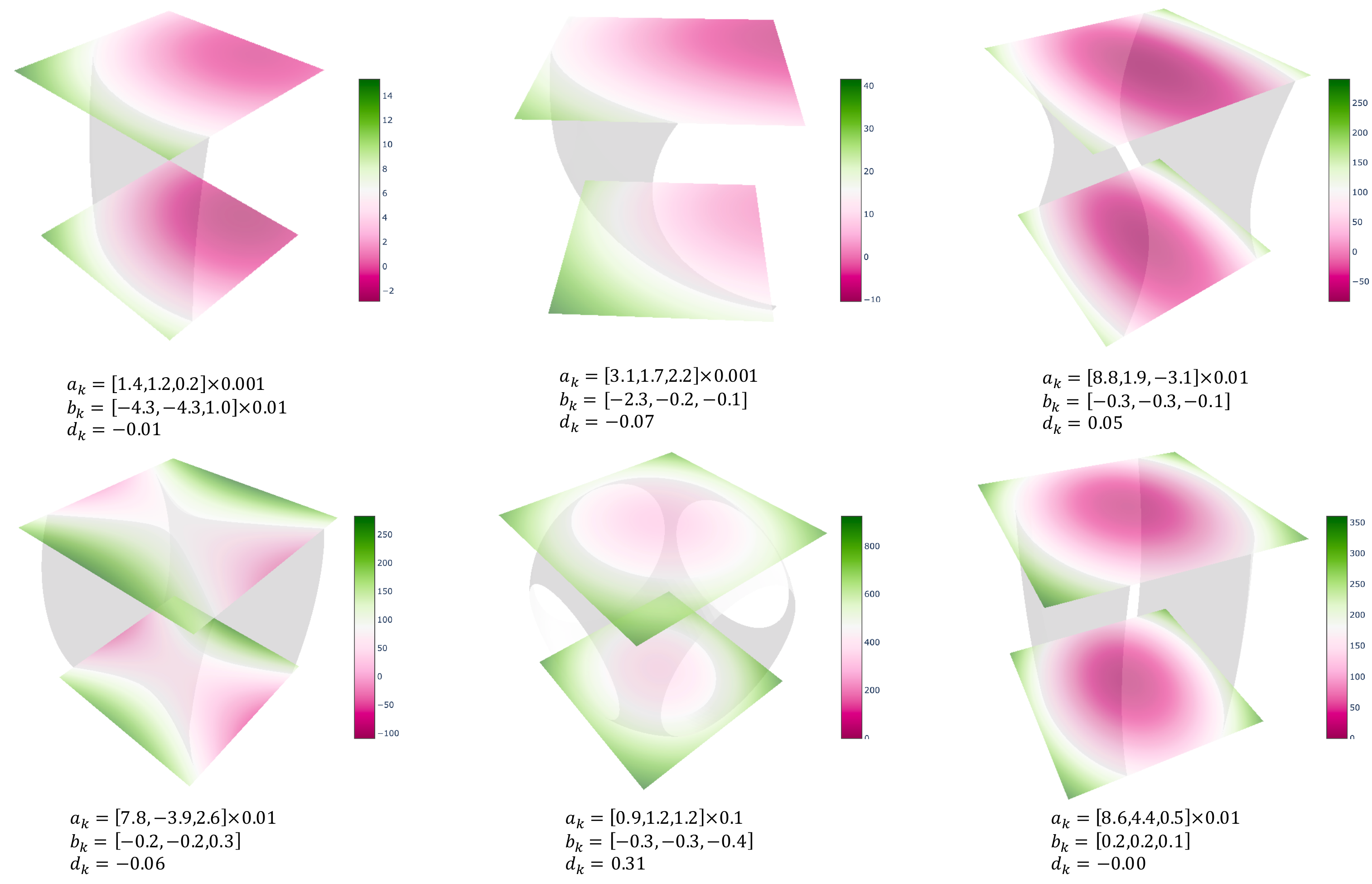}
    \caption{The visualization of learned quadratic potential fields, which are randomly selected from different layers. The gray surface is defined by $\{\mathbf{y}_i | \tilde{p}_k(\mathbf{y}_i) = \eta, \eta \in \mathbb{R}^+\}$.}
    \label{fig:pot}
\end{figure*}

Here, we provide an analysis from the geometric aspect to explain why our proposed linear and quadratic potential function works better than existing point convolution methods. 

As for $\tilde{p}_k^1$, we can recognize the linear terms as a point-normal equation of plane. Therefore, the linear potential fields shape planes lied in different directions on local space, \ie $\mathcal{B}_{r}^3$. 
The potential value of input point indicates different distance to planes. 
If we take normal vector $\mathbf{n}_{\mathbf{y_i}}$ into account, $\tilde{p}_k^\prime$ also encodes the direction between the plane and input points.
In order to better demonstrate the learned linear potential fields with different directions, we colorize each point with the color of most related potential fields. 
As shown in Fig.~\ref{fig:plane}, we find that points lie in different positions activate with respect to different potential fields. 

As for $\tilde{p}_k^2$, it is a special kind of quadratic surface equation with removing the cross items. However, this simplified formulation can still maintain a lot of different geometric shape. Every neighborhood $\mathcal{N}_x$ can build a local shape, and only the local shape matches well with the quadratic surface, they can have a stronger correlation value. That is very different from discrete convolution, which only takes the point-to-point relationship into consideration. In contrast, our potential convolution will take the point-to-shape relationship into account. 
We visualize the learned quadratic potential fields in Fig.~\ref{fig:pot}. 

In conclusion, we claim that the linear potential function can well encode the direction and distance information of input points, and the quadratic potential function can better encode more complex shape information between local neighborhoods and potential fields.

\subsection{Simplification}
In previous sections, we describe the computation process on only one single potential field.
The naive implementation of $\tilde{p}_k^1, \tilde{p}_k^\prime, \tilde{p}_k^2$ is inefficient, and we find that the computation process of linear and quadratic functions can be highly paralleled by formulating them as point-wise convolution operations.

We can rewrite the above Eq.~\ref{eq:pk1}, Eq.~\ref{eq:pk'}, Eq.~\ref{eq:pk2} to paralleled matrix multiplication:
\begin{equation}
    \begin{split}
    \tilde{P}^1(\mathbf{y}_i)&=A \mathbf{y}_i^T+D \\
    \tilde{P}^\prime (\mathbf{y}_i, \mathbf{n}_{\mathbf{y}_i})&=A(\mathbf{y}_i+\mathbf{n}_{\mathbf{y}_i})^T + D \\
    \tilde{P}^2(\mathbf{y}_i) &= A \mathbf{y}_i^T + B (\mathbf{y}_i^2)^T + D,
    \end{split}
    \label{eq:parallel}
\end{equation}
where $A \in \mathbb{R}^{D^\prime \times 3}, B \in \mathbb{R}^{D^\prime\times 3}, D \in \mathbb{R}^{D^\prime}$ are the stacked parameters of potential function, \ie $A_{k:}=\mathbf{a}_k, B_{k:}=\mathbf{b}_k, D_{k}=d_k$.

And then, we can obtain the convolution kernels by:
\begin{equation}
    g(\mathbf{y_i})=(h \circ \tilde{P})(\mathbf{y}_i) W,
\label{eq:g}
\end{equation}
where $W \in \mathbb{R}^{D^\prime\times D}$ and $W_{k:}=w_k$.
Then, we put Eq.~\ref{eq:g} into Eq.~\ref{eq:pointconv}, yields:
\begin{equation}
    (\mathcal{F} \ast g)(\mathbf{x}) = \sum_{\mathbf{x}_i \in \mathcal{N}_\mathbf{x}} (h \circ \tilde{P})(\mathbf{x}_i-\mathbf{x}) W\mathbf{f}_i.
\end{equation}

\paragraph{Implementation Details}
The above formulation shows that potential convolution can be formulated as two terms, the potential fields related term $(h \circ \tilde{P})(\mathbf{x}_i-\mathbf{x})$, and the feature related term $W\mathbf{f}_i$.
As for $(h \circ \tilde{P})(\mathbf{x}_i-\mathbf{x})$, the main computation is brought by the calculation of potential value. As shown in Eq.~\ref{eq:parallel}, we can implement each matrix multiplication as a point-wise convolution operator. 
As for $\tilde{P}^1, \tilde{P}^\prime$, we can implement it via a point-wise convolution, whose kernel weight is $A$ and bias is $D$.
As for $\tilde{P}^2$, we can implement it via two point-wise convolution operators, one of them contains the kernel weight $A$ without bias, and the other contains the kernel weight $B$ with bias $D$.
Sequentially, $W\mathbf{f}_i$ can be implemented via a point-wise convolution, whose kernel weight is $W$.

\section{Experiments}
\begin{figure*}
    \centering
    \includegraphics[width=0.8\textwidth]{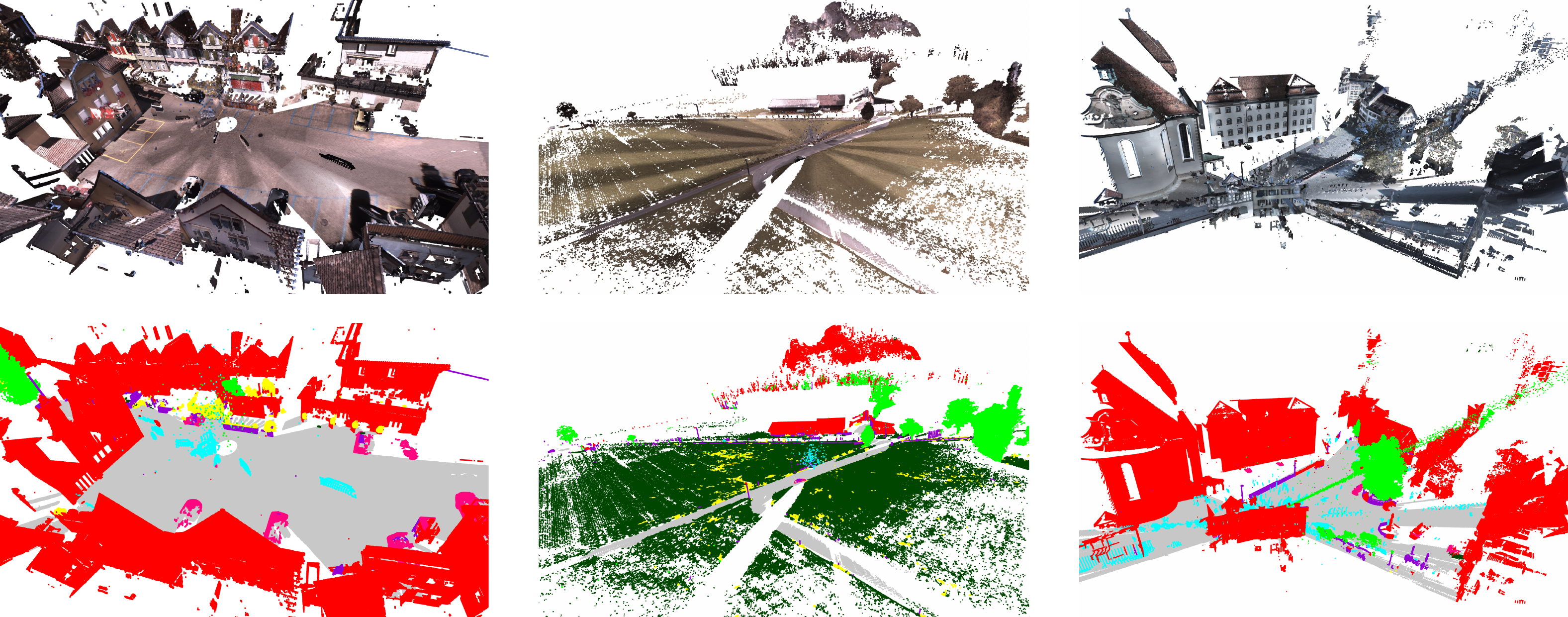}
    \caption{Segmentation results of Semantic3D. We colorize each point with corresponding labels’ color. The first row is input color point clouds, and the second one is segmentation results.}
    \label{fig:sem3d}
\end{figure*}

\begin{table*}[tbp]
\centering
\small
\resizebox{\textwidth}{!}{
\begin{tabular}{l|ccc|ccccccccccccc} 
    Methods                                   & OA(\%) & mAcc(\%)  & mIoU(\%) & ceil.  & floor & wall & beam  & col. & wind.  & door & chair  & table  & book.  & sofa & board  & clut.  \\ \hline
    PointNet~\cite{qi2017pointnet}             & -      & 49.0      & 41.1     & 88.8   & 97.3  & 69.8 & 0.1   & 3.9  & 46.3   & 10.8 & 52.6   & 58.9   & 40.3   & 5.9  & 26.4   & 33.2 \\
    PointCNN~\cite{li2018pointcnn}             & 85.9   & 63.9      & 57.3     & 92.3   & 98.2  & 79.4 & 0.0   & 17.6 & 22.8   & 62.1 & 80.6   & 74.4   & 66.7   & 31.7 & 62.2   & 56.7 \\
    TangentConv~\cite{tatarchenko2018tangent}  & -      & 62.2      & 52.6     & 90.5   & 97.7  & 74.0 & 0.0   & 20.7 & 39.0   & 31.3 & 69.4   & 77.5   & 38.5   & 57.3 & 48.8   & 39.8 \\
    ParamConv~\cite{wang2018deep}              & -      & 67.1      & 58.3     & 92.3   & 96.2  & 75.9 & \textbf{0.3} & 6.0  & \textbf{69.5} & 63.5 & 66.9 & 65.6 & 47.3 & 68.9 & 59.1 & 46.2 \\
    SegCloud~\cite{tchapmi2017segcloud}        & -      & 57.4      & 48.9     & 90.1   & 96.1  & 69.9 & 0.0 & 18.4 & 38.4 & 23.1 & 75.9 & 70.4 & 58.4 & 40.9 & 13.0 & 41.6 \\
    SPGraph~\cite{landrieu2018large}           & 86.4   & 66.5      & 58.0     & 89.4   & 96.9  & 78.1 & 0.0 & \textbf{42.8} & 48.9 & 61.6 & 84.7 & 75.4 & 69.8 & 52.6 & 2.1  & 52.2 \\
    Eff 3D Conv~\cite{zhang2018efficient}      & -      & 68.3      & 51.8     & 79.8   & 93.9  & 69.0 & 0.2 & 28.3 & 38.5 & 48.3 & 71.1 & 73.6 & 48.7 & 59.2 & 29.3 & 33.1 \\
    RNN Fusion~\cite{ye20183d}                 & -      & 63.9      & 57.3     & 92.3   & 98.2  & 79.4 & 0.0 & 17.6 & 22.8 & 62.1 & 74.4 & 80.6 & 31.7 & 66.7 & 62.1 & 56.7 \\
    SSP+SPG~\cite{landrieu2019point}           & 87.9   & 68.2      & 61.7     & 91.9   & 96.7  & 80.8 & 0.0 & 28.8 & 60.3 & 57.2 & 85.5 & 76.4 & 70.5 & 49.1 & 51.6 & 53.3 \\ \hline
    \multicolumn{17}{c}{Point Convolution} \\ \hline
    KPConv~\cite{thomas2019kpconv}             & -      & 70.9      & 65.4     & 92.6   & 97.3  & 81.4 & 0.0 & 16.5 & 54.5 & \textbf{69.5} & 90.1 & 80.2 & 74.6 & 66.4 & 63.7 & 58.1 \\
    FPConv~\cite{lin2020fpconv}                & -      & -         & 62.8     & 94.6   & \textbf{98.5} & 80.9 & 0.0 & 19.1 & 60.1 & 48.9 & 80.6 & \textbf{88.0} & 53.2 & 68.4 & 68.2 & 54.9 \\ \hline
    SSP+SPG$\bullet \tilde{P}^\prime$        & 87.7   & 70.4      & 62.1     & 92.6   & 96.6  & 81.6 & 0.0 & 36.3 & 54.9 & 50.3 & 84.8 & 77.0 & 70.7 & 60.9 & 47.7 & 53.2 \\
    KPConv$\circ \tilde{P}^2$           & 89.9 & 72.8 & \textbf{67.7}     & \textbf{94.7} & \textbf{98.5} & 82.4 & 0.0 & 28.6 & 55.9 & 67.3 & \textbf{91.6} & 82.8 & 74.7 & \textbf{74.0} & \textbf{68.7} & \textbf{60.4} \\
    KPConv$\circ \tilde{P}^m$       & \textbf{90.0} & \textbf{73.7} & \textbf{67.7} & 94.3 & \textbf{98.5} & \textbf{83.3} & 0.0 & 33.9 & 57.9 & 65.8 & 90.4 & 82.5 & \textbf{74.9} & 70.5 & 68.5 & 58.9
\end{tabular}
}
\caption{Semantic segmentation results on S3DIS. $\bullet$ indicates that we reuse the previous network frameworks but replace the original point-wise convolution operation with our potential convolution. $\circ$ indicates that we plugin our potential convolution operation to the previous network frameworks. We list the most competitive point convolution methods, some others also reported as a reference.}
\label{tab:s3dis}
\end{table*}

In order to evaluate the performance of potential convolution, we conduct experiments on several widely used datasets, ModelNet40~\cite{wu20153d}, ShapeNet~\cite{yi2016scalable}, Semantic3D~\cite{hackel2017semantic3d} and S3DIS~\cite{armeni20163d}. 
To more fairly compare the effectiveness of our methods, we do not attempt to design a special network structure for potential convolution, but reuse the popular networks proposed by other related methods and plug in our potential convolution to the basic framework or replace the original point convolution with ours. We also follow the default experiment settings as original papers and do not select specified hyper-parameters for potential convolution. All experiments show that potential convolution can well adapt to different structures and is not sensitive to experiment settings.
\subsection{3D Shape Classification}
We use ModelNet40~\cite{wu20153d} to evaluate the classification performance of our methods.
There are 12,311 CAD models from 40 man-made object categories, split into 9,843 for training and 2,468 for testing.
Rather than designing new network models, we directly reuse the most simple and popular model from PointNet~\cite{qi2017pointnet} and PointNet++~\cite{qi2017pointnet++} but replace the original point-wise convolution with our potential convolution. 

\paragraph{Classification Performance on ModelNet40}
\begin{table}[tbp]
\centering
\small
\begin{tabular}{lcc}
    Methods                                 & OA(\%)        & Settings       \\ \hline
    Subvolume~\cite{qi2016volumetric}        & 89.2          & voxel          \\
    MVCNN~\cite{hermosilla2018monte}         & 92.0          & mesh           \\
    PointNet~\cite{qi2017pointnet}           & 89.2          & 1024 pts       \\
    PointNet++~\cite{qi2017pointnet++}       & 91.9          & 5000 pts       \\
    ECC~\cite{simonovsky2017dynamic}         & 87.4          & graph          \\ \hline
    \multicolumn{3}{c}{Point Convolution}                                    \\ \hline
    KPConv~\cite{thomas2019kpconv}           & 92.9          & 6800 avg. pts  \\
    FPConv~\cite{lin2020fpconv}              & 92.5          & 1024 pts + nor \\
    PointConv~\cite{wu2019pointconv}         & 92.5          & 1024 pts + nor \\
    ConvPoint~\cite{boulch2020convpoint}     & 92.5          & 2048 pts + nor \\ \hline
    PointNet$\bullet \tilde{P}^1$            & 91.5          & 1024 pts       \\
    PointNet$\bullet \tilde{P}^\prime$            & 92.6          & 1024 pts + nor \\
    PointNet$\bullet \tilde{P}^2$            & 91.6          & 1024 pts       \\
    PointNet++$\bullet \tilde{P}^\prime$          & \textbf{93.0} & 1024 pts + nor
\end{tabular}
\caption{Classification accuracy on ModelNet40. $\bullet$ indicates that we reuse the previous network frameworks but replace the original point-wise convolution operation with our potential convolution.}
\label{tab:modelnet40}
\end{table}
As shown in Tab.~\ref{tab:modelnet40}, our potential convolution outperforms other methods. By simply replacing the original point-wise convolution of PointNet~\cite{qi2017pointnet}, the performance has been largely improved. What's more, normal vector gives benefit to the final accuracy, too. However, the quadratic potential convolution seems more proven to be overfitting in such an easy task, and achieves a lower performance in the validation set.

\paragraph{Tolerance of sparse point clouds}
\begin{table}[tbp]
\centering
\small
\begin{tabular}{l|ccccc}
    \# Points                           & 64            & 128           & 256           & 512           & 1024          \\ \hline
    PointConv~\cite{wu2019pointconv}    & 86.8          & 87.8          & 89.9          & 90.8          & 92.5          \\
    KPConv~\cite{thomas2019kpconv}      & 75.3          & 81.8          & 86.3          & 88.5          & 90.0          \\ \hline
    Ours                                & \textbf{90.2} & \textbf{91.1} & \textbf{91.5} & \textbf{91.6} & \textbf{92.6}
    \end{tabular}
\caption{The performance under sparse input. (OA (\%) is reported.)}
\label{tab:sparse}
\end{table}

\begin{table*}[tbp]
\centering
\small
\resizebox{\textwidth}{!}{
\begin{tabular}{l|cc|cccccccc}
    Methods      & OA(\%)        & mIoU(\%) & man-made.     & natural.      & high veg.     & low veg.      & buildings     & hard scape    & scanning art. & cars          \\ \hline
    TMLC-MSR~\cite{hackel2016fast}                & 86.2          & 54.2     & 89.8          & 74.5          & 53.7          & 26.8          & 88.8          & 18.9          & 36.4          & 44.7          \\
    DeePr3SS~\cite{lawin2017deep}                 & 88.9          & 58.5     & 85.6          & 83.2          & 74.2          & 32.4          & 89.7          & 18.5          & 25.1          & 59.2          \\
    SnapNet~\cite{boulch2017unstructured}         & 88.6          & 59.1     & 82.0          & 77.3          & 79.7          & 22.9          & 91.1          & 18.4          & 37.3          & 64.4          \\
    RF\_MSSF~\cite{thomas2018semantic}            & 90.3          & 62.7     & 87.6          & 80.3          & 81.8          & 36.4          & 92.2          & 24.1          & 42.6          & 56.6          \\
    MSDeepVoxNet~\cite{roynard2018classification} & 88.4          & 65.3     & 83.0          & 67.2          & 83.8          & 36.7          & 92.4          & 31.3          & 50.0          & 78.2          \\
    ShellNet~\cite{zhang2019shellnet}             & 93.2          & 69.3     & 96.3          & 90.4          & 83.9          & 41.0          & 94.2          & 34.7          & 43.9          & 70.2          \\
    SegCloud~\cite{tchapmi2017segcloud}           & 88.1          & 61.3     & 83.9          & 66.0          & 86.0          & 40.5          & 91.1          & 30.9          & 27.5          & 64.3          \\
    SPGraph~\cite{landrieu2018large}              & 94.0          & 73.2     & 97.4          & \textbf{92.6} & \textbf{87.9} & 44.0          & 93.2          & 31.0          & 63.5          & 76.2          \\ \hline
    \multicolumn{11}{c}{Point Convolution}    \\ \hline
    KPConv~\cite{thomas2019kpconv}      & 92.9          & 74.6     & 90.9          & 82.2          & 84.2          & 47.9          & \textbf{94.9} & 40.0          & \textbf{77.3} & \textbf{79.7} \\
    SPGraph$\bullet \tilde{P}^2$   & \textbf{94.4} & \textbf{74.8}     & \textbf{97.5} & 90.7          & 86.8          & \textbf{48.4} & 94.6          & \textbf{41.2} & 66.8          & 72.7
\end{tabular}
}
\caption{Semantic segmentation results on Semantic3D. $\bullet$ indicates that we reuse the previous network frameworks but replace the original point-wise convolution operation with our potential convolution. We list the most competitive point convolution methods, some others also reported as a reference.}
\label{tab:sem3d}
\end{table*}

In this experiment, we compare the performance of different kernels under sparse input.
We select KPConv~\cite{thomas2019kpconv} and PointConv~\cite{wu2019pointconv}, the most competitive discrete and continuous convolution methods, as our baseline.
We use the same network as PointNet~\cite{qi2017pointnet} but replace the point-wise convolution with our potential convolution.

All the experiments are implemented using the PyTorch~\cite{paszke2017automatic} deep leaning framework. 
The implementation of KPConv~\footnote{\url{https://github.com/HuguesTHOMAS/KPConv-PyTorch}} and PointConv~\footnote{\url{https://github.com/DylanWusee/pointconv_pytorch}} are both directly taken from the official released source code.
All experiment settings keep the same with the default implementation, except that the input point number is changed. 
We also use the corresponding protocols to train and evaluate with default parameters, which means that the training and evaluation phase may vary among different methods. We think it would be fair to compare performance for different methods with their original settings.

As shown in Tab.~\ref{tab:sparse}, either discrete or continuous convolution suffers from the sparse input points, but our potential convolution has a better tolerance to this issue, and can even achieve an accuracy over 90\% with only 64 points as input.

\subsection{3D Shape Segmentation}

\begin{table}[tbp]
\small
\centering
\begin{tabular}{l|cc}
    Methods                             & OA(\%) & mAcc(\%) \\ \hline
    Kd-Net~\cite{klokov2017escape}      & 77.4   & 82.3    \\
    SO-Net~\cite{li2018so}              & 81.0   & 84.9    \\
    PCNN by Ext~\cite{atzmon2018point}  & 81.8   & 85.1    \\
    PointNet++~\cite{qi2017pointnet++}  & 81.9   & 85.1    \\
    DGCNN~\cite{wang2019dynamic}        & 82.3   & 85.1    \\
    SPLATNet~\cite{su2018splatnet}      & 83.7   & 85.4    \\ \hline
    \multicolumn{3}{c}{Point Convolution}  \\ \hline
    KPConv~\cite{thomas2019kpconv}      & 85.0   & 86.2    \\
    SynSpecCNN~\cite{yi2017syncspeccnn} & 82.0   & 84.7    \\
    Spidercnn~\cite{xu2018spidercnn}    & 82.4   & 85.3    \\
    SubSparseCNN~\cite{graham20183d}    & 83.3   & 86.0    \\
    PointCNN~\cite{li2018pointcnn}      & 84.6   & 86.1    \\
    FlexConv~\cite{groh2018flex}        & 85.0   & 84.7    \\ \hline
    PointNet++$\bullet \tilde{P}\prime$ & 82.0   & 85.4    \\
    KPConv$\circ \tilde{P}^2$           & \textbf{85.1}   & \textbf{86.3}    \\
\end{tabular}
\caption{Results on ShapeNet part dataset. Class avg. is the mean IoU averaged across all object categories, and instance avg. is the mean IoU across all objects. $\circ$ indicates that we plugin our potential convolution operation to the previous network frameworks.}
\label{tab:shapenet_short}
\end{table}

Given a point cloud, the part segmentation task is to recognize the different constitutive parts of the shape.
We evaluate our model on the ShapeNet~\cite{yi2016scalable} part segmentation benchmark. It is composed of 16680 models belonging to 16 shape categories and split train/test sets. Each category is annotated with 2-to-6-part labels, from 50-part classes in total.

We use point class and instance average intersection-over-union (IoU) to evaluate our potential convolution, same as other part segmentation algorithms~\cite{qi2017pointnet, qi2017pointnet++, thomas2019kpconv, wu2019pointconv, boulch2020convpoint}. As shown in Tab.~\ref{tab:shapenet_short}, we have achieved better performance than baseline and achieve comparable results among all methods.
\begin{figure}
    \centering
    \includegraphics[width=3in]{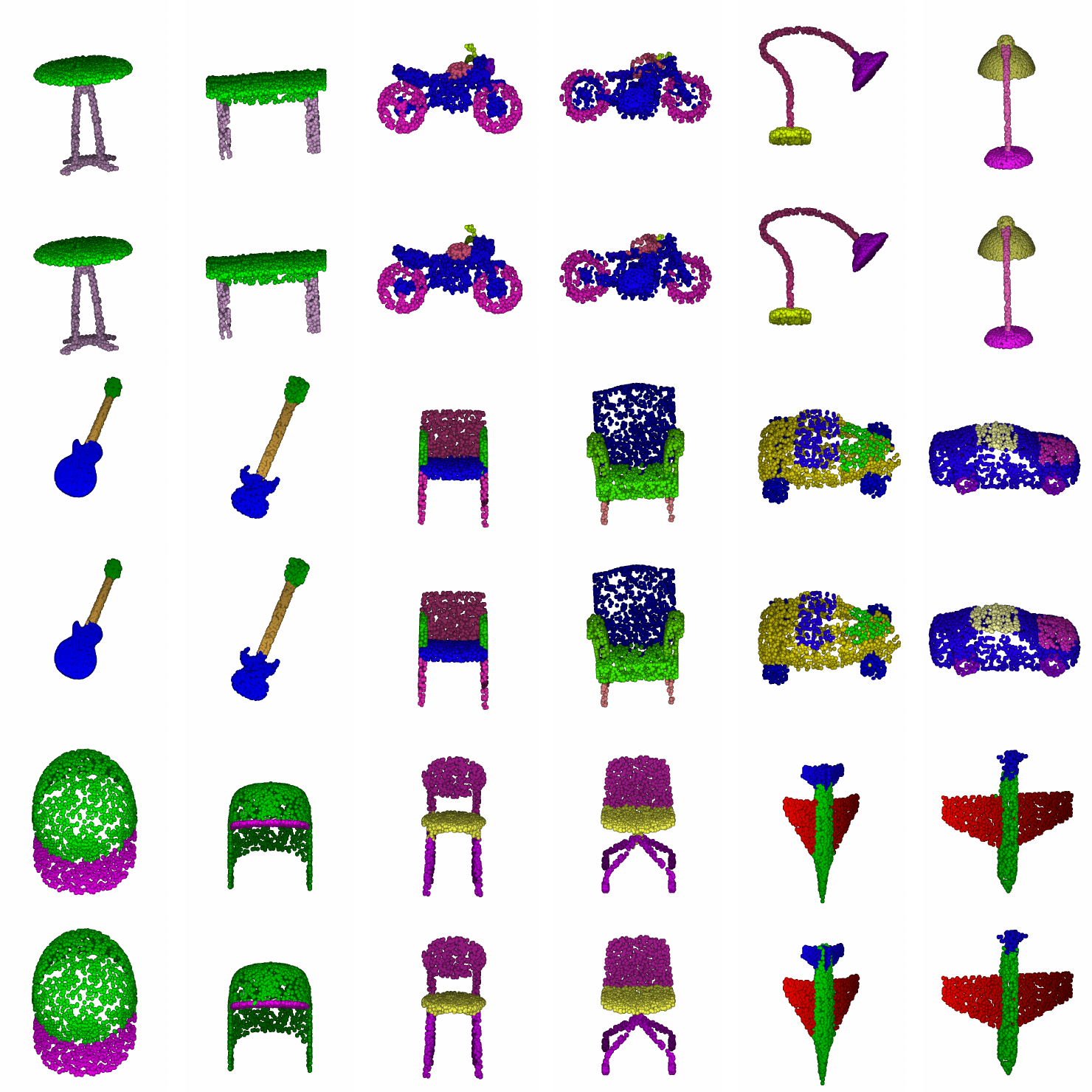}
    \caption{Part segmentation results. For each pair of objects, the upper one is the ground truth, the lower one is predicted by our method. Best viewed in color.
    }
    \label{fig:shapenet}
\end{figure}

We also visualize the segmentation results in Fig.~\ref{fig:shapenet}. 

\subsection{3D Scene Segmentation}
This experiment shows how potential convolution generalizes to indoor and outdoor scenes. To this end, we choose to evaluate our method on the indoor S3DIS~\cite{armeni20163d} dataset and outdoor Semantic3D~\cite{hackel2017semantic3d} datasets.
S3DIS covers six large-scale indoor areas from three different buildings with a total of 273 million points annotated with 13 classes. Like ~\cite{tchapmi2017segcloud}, we advocate the use of Area-5 as the test scene to better measure the generalization ability of our method. Semantic3D is an online benchmark comprising several fixed LiDAR scans of different outdoor scenes. More than 4 billion points are annotated with 8 classes in this dataset, but they mostly cover ground, building or vegetation and there are fewer object instances than other datasets. We favor the reduced-8 challenge because it is less biased by the objects close to the scanner~\cite{thomas2019kpconv}.

\paragraph{Performance on S3DIS}

As shown in Tab.~\ref{tab:s3dis}, our methods can achieve state of the art performance in indoor S3DIS scenes. Furthermore, we also devise a potential function $\tilde{P}^{m}$ with a three-layers point-wise convolution operation with $(3, 8, D^\prime)$ hidden size. It is obvious that with a more complex potential function, the performance can be further improved.

\paragraph{Performance on Semantic3D}

As shown in Tab.~\ref{tab:sem3d}, our methods can achieve state of the art performance in outdoor Semantic3D scenes. We visualize the segmentation results on Semantic3D in Fig.~\ref{fig:sem3d}.
\section{Conclusion}
% In this work, we propose a novel approach to perform a convolution operation on 3D point clouds, named potential convolution.
% With the simplest design of potential function, \ie. linear and quadratic ones, we achieve the state-of-the-art performance under shape classification and indoor/outdoor scene segmentation. 
% Furthermore, potential convolution can be efficiently implemented with little computation overhead. We believe that a dedicate design of various potential functions will further improve the performance largely, and we leave it as a feature work.

In this work, we propose a new type of point convolution, named as \textit{potential convolution}. A learnable potential field is associated with each kernel to undercover the relationship between the points and the kernel weight. We provide two simple yet effective potential functions, explain why they would work and how they could be implemented efficiently using pointwise convolutions. Thorough experiments on the 3D vision tasks including shape classification, shape segmentation and scene segmentation demonstrate the superior performance of our methods.   

{\small
\bibliographystyle{ieee_fullname}
\bibliography{egbib}

\begin{thebibliography}{10}\itemsep=-1pt

\bibitem{armeni20163d}
Iro Armeni, Ozan Sener, Amir~R Zamir, Helen Jiang, Ioannis Brilakis, Martin
  Fischer, and Silvio Savarese.
\newblock 3d semantic parsing of large-scale indoor spaces.
\newblock In {\em Proceedings of the IEEE Conference on Computer Vision and
  Pattern Recognition}, pages 1534--1543, 2016.

\bibitem{atzmon2018point}
Matan Atzmon, Haggai Maron, and Yaron Lipman.
\newblock Point convolutional neural networks by extension operators.
\newblock {\em arXiv preprint arXiv:1803.10091}, 2018.

\bibitem{ben20183dmfv}
Yizhak Ben-Shabat, Michael Lindenbaum, and Anath Fischer.
\newblock 3dmfv: Three-dimensional point cloud classification in real-time
  using convolutional neural networks.
\newblock {\em IEEE Robotics and Automation Letters}, 3(4):3145--3152, 2018.

\bibitem{boulch2020convpoint}
Alexandre Boulch.
\newblock Convpoint: Continuous convolutions for point cloud processing.
\newblock {\em Computers \& Graphics}, 2020.

\bibitem{boulch2017unstructured}
Alexandre Boulch, Bertrand Le~Saux, and Nicolas Audebert.
\newblock Unstructured point cloud semantic labeling using deep segmentation
  networks.
\newblock {\em 3DOR}, 2:7, 2017.

\bibitem{bounini2017modified}
Farid Bounini, Denis Gingras, Herve Pollart, and Dominique Gruyer.
\newblock Modified artificial potential field method for online path planning
  applications.
\newblock In {\em 2017 IEEE Intelligent Vehicles Symposium (IV)}, pages
  180--185. IEEE, 2017.

\bibitem{dai2017deformable}
Jifeng Dai, Haozhi Qi, Yuwen Xiong, Yi Li, Guodong Zhang, Han Hu, and Yichen
  Wei.
\newblock Deformable convolutional networks.
\newblock In {\em Proceedings of the IEEE international conference on computer
  vision}, pages 764--773, 2017.

\bibitem{graham20183d}
Benjamin Graham, Martin Engelcke, and Laurens Van Der~Maaten.
\newblock 3d semantic segmentation with submanifold sparse convolutional
  networks.
\newblock In {\em Proceedings of the IEEE conference on computer vision and
  pattern recognition}, pages 9224--9232, 2018.

\bibitem{graham2017submanifold}
Benjamin Graham and Laurens van~der Maaten.
\newblock Submanifold sparse convolutional networks.
\newblock {\em arXiv preprint arXiv:1706.01307}, 2017.

\bibitem{groh2018flex}
Fabian Groh, Patrick Wieschollek, and Hendrik~PA Lensch.
\newblock Flex-convolution (deep learning beyond grid-worlds).
\newblock {\em arXiv preprint arXiv:1803.07289}, 2, 2018.

\bibitem{hackel2017semantic3d}
Timo Hackel, Nikolay Savinov, Lubor Ladicky, Jan~D Wegner, Konrad Schindler,
  and Marc Pollefeys.
\newblock Semantic3d. net: A new large-scale point cloud classification
  benchmark.
\newblock {\em arXiv preprint arXiv:1704.03847}, 2017.

\bibitem{hackel2016fast}
Timo Hackel, Jan~D Wegner, and Konrad Schindler.
\newblock Fast semantic segmentation of 3d point clouds with strongly varying
  density.
\newblock {\em ISPRS annals of the photogrammetry, remote sensing and spatial
  information sciences}, 3:177--184, 2016.

\bibitem{hermosilla2018monte}
Pedro Hermosilla, Tobias Ritschel, Pere-Pau V{\'a}zquez, {\`A}lvar Vinacua, and
  Timo Ropinski.
\newblock Monte carlo convolution for learning on non-uniformly sampled point
  clouds.
\newblock {\em ACM Transactions on Graphics (TOG)}, 37(6):1--12, 2018.

\bibitem{hua2018pointwise}
Binh-Son Hua, Minh-Khoi Tran, and Sai-Kit Yeung.
\newblock Pointwise convolutional neural networks.
\newblock In {\em Proceedings of the IEEE Conference on Computer Vision and
  Pattern Recognition}, pages 984--993, 2018.

\bibitem{klokov2017escape}
Roman Klokov and Victor Lempitsky.
\newblock Escape from cells: Deep kd-networks for the recognition of 3d point
  cloud models.
\newblock In {\em Proceedings of the IEEE International Conference on Computer
  Vision}, pages 863--872, 2017.

\bibitem{landrieu2019point}
Loic Landrieu and Mohamed Boussaha.
\newblock Point cloud oversegmentation with graph-structured deep metric
  learning.
\newblock In {\em Proceedings of the IEEE Conference on Computer Vision and
  Pattern Recognition}, pages 7440--7449, 2019.

\bibitem{landrieu2018large}
Loic Landrieu and Martin Simonovsky.
\newblock Large-scale point cloud semantic segmentation with superpoint graphs.
\newblock In {\em Proceedings of the IEEE Conference on Computer Vision and
  Pattern Recognition}, pages 4558--4567, 2018.

\bibitem{lawin2017deep}
Felix~J{\"a}remo Lawin, Martin Danelljan, Patrik Tosteberg, Goutam Bhat,
  Fahad~Shahbaz Khan, and Michael Felsberg.
\newblock Deep projective 3d semantic segmentation.
\newblock In {\em International Conference on Computer Analysis of Images and
  Patterns}, pages 95--107. Springer, 2017.

\bibitem{li2018so}
Jiaxin Li, Ben~M Chen, and Gim Hee~Lee.
\newblock So-net: Self-organizing network for point cloud analysis.
\newblock In {\em Proceedings of the IEEE conference on computer vision and
  pattern recognition}, pages 9397--9406, 2018.

\bibitem{li2018pointcnn}
Yangyan Li, Rui Bu, Mingchao Sun, Wei Wu, Xinhan Di, and Baoquan Chen.
\newblock Pointcnn: Convolution on x-transformed points.
\newblock In {\em Advances in neural information processing systems}, pages
  820--830, 2018.

\bibitem{li2016fpnn}
Yangyan Li, Soeren Pirk, Hao Su, Charles~R Qi, and Leonidas~J Guibas.
\newblock Fpnn: Field probing neural networks for 3d data.
\newblock In {\em Advances in Neural Information Processing Systems}, pages
  307--315, 2016.

\bibitem{lin2020fpconv}
Yiqun Lin, Zizheng Yan, Haibin Huang, Dong Du, Ligang Liu, Shuguang Cui, and
  Xiaoguang Han.
\newblock Fpconv: Learning local flattening for point convolution.
\newblock In {\em Proceedings of the IEEE/CVF Conference on Computer Vision and
  Pattern Recognition}, pages 4293--4302, 2020.

\bibitem{maturana2015voxnet}
Daniel Maturana and Sebastian Scherer.
\newblock Voxnet: A 3d convolutional neural network for real-time object
  recognition.
\newblock In {\em 2015 IEEE/RSJ International Conference on Intelligent Robots
  and Systems (IROS)}, pages 922--928. IEEE, 2015.

\bibitem{park2001obstacle}
Min~Gyu Park, Jae~Hyun Jeon, and Min~Cheol Lee.
\newblock Obstacle avoidance for mobile robots using artificial potential field
  approach with simulated annealing.
\newblock In {\em ISIE 2001. 2001 IEEE International Symposium on Industrial
  Electronics Proceedings (Cat. No. 01TH8570)}, volume~3, pages 1530--1535.
  IEEE, 2001.

\bibitem{paszke2017automatic}
Adam Paszke, Sam Gross, Soumith Chintala, Gregory Chanan, Edward Yang, Zachary
  DeVito, Zeming Lin, Alban Desmaison, Luca Antiga, and Adam Lerer.
\newblock Automatic differentiation in pytorch.
\newblock 2017.

\bibitem{qi2017pointnet}
Charles~R Qi, Hao Su, Kaichun Mo, and Leonidas~J Guibas.
\newblock Pointnet: Deep learning on point sets for 3d classification and
  segmentation.
\newblock In {\em Proceedings of the IEEE conference on computer vision and
  pattern recognition}, pages 652--660, 2017.

\bibitem{qi2016volumetric}
Charles~R Qi, Hao Su, Matthias Nie{\ss}ner, Angela Dai, Mengyuan Yan, and
  Leonidas~J Guibas.
\newblock Volumetric and multi-view cnns for object classification on 3d data.
\newblock In {\em Proceedings of the IEEE conference on computer vision and
  pattern recognition}, pages 5648--5656, 2016.

\bibitem{qi2017pointnet++}
Charles~Ruizhongtai Qi, Li Yi, Hao Su, and Leonidas~J Guibas.
\newblock Pointnet++: Deep hierarchical feature learning on point sets in a
  metric space.
\newblock In {\em Advances in neural information processing systems}, pages
  5099--5108, 2017.

\bibitem{ravanbakhsh2016deep}
Siamak Ravanbakhsh, Jeff Schneider, and Barnabas Poczos.
\newblock Deep learning with sets and point clouds.
\newblock {\em arXiv preprint arXiv:1611.04500}, 2016.

\bibitem{riegler2017octnet}
Gernot Riegler, Ali Osman~Ulusoy, and Andreas Geiger.
\newblock Octnet: Learning deep 3d representations at high resolutions.
\newblock In {\em Proceedings of the IEEE Conference on Computer Vision and
  Pattern Recognition}, pages 3577--3586, 2017.

\bibitem{roynard2018classification}
Xavier Roynard, Jean-Emmanuel Deschaud, and Fran{\c{c}}ois Goulette.
\newblock Classification of point cloud scenes with multiscale voxel deep
  network.
\newblock {\em arXiv preprint arXiv:1804.03583}, 2018.

\bibitem{simonovsky2017dynamic}
Martin Simonovsky and Nikos Komodakis.
\newblock Dynamic edge-conditioned filters in convolutional neural networks on
  graphs.
\newblock In {\em Proceedings of the IEEE conference on computer vision and
  pattern recognition}, pages 3693--3702, 2017.

\bibitem{su2018splatnet}
Hang Su, Varun Jampani, Deqing Sun, Subhransu Maji, Evangelos Kalogerakis,
  Ming-Hsuan Yang, and Jan Kautz.
\newblock Splatnet: Sparse lattice networks for point cloud processing.
\newblock In {\em Proceedings of the IEEE Conference on Computer Vision and
  Pattern Recognition}, pages 2530--2539, 2018.

\bibitem{tatarchenko2018tangent}
Maxim Tatarchenko, Jaesik Park, Vladlen Koltun, and Qian-Yi Zhou.
\newblock Tangent convolutions for dense prediction in 3d.
\newblock In {\em Proceedings of the IEEE Conference on Computer Vision and
  Pattern Recognition}, pages 3887--3896, 2018.

\bibitem{tchapmi2017segcloud}
Lyne Tchapmi, Christopher Choy, Iro Armeni, JunYoung Gwak, and Silvio Savarese.
\newblock Segcloud: Semantic segmentation of 3d point clouds.
\newblock In {\em 2017 international conference on 3D vision (3DV)}, pages
  537--547. IEEE, 2017.

\bibitem{thomas2018semantic}
Hugues Thomas, Fran{\c{c}}ois Goulette, Jean-Emmanuel Deschaud, Beatriz
  Marcotegui, and Yann LeGall.
\newblock Semantic classification of 3d point clouds with multiscale spherical
  neighborhoods.
\newblock In {\em 2018 International conference on 3D vision (3DV)}, pages
  390--398. IEEE, 2018.

\bibitem{thomas2019kpconv}
Hugues Thomas, Charles~R Qi, Jean-Emmanuel Deschaud, Beatriz Marcotegui,
  Fran{\c{c}}ois Goulette, and Leonidas~J Guibas.
\newblock Kpconv: Flexible and deformable convolution for point clouds.
\newblock In {\em Proceedings of the IEEE International Conference on Computer
  Vision}, pages 6411--6420, 2019.

\bibitem{vadakkepat2000evolutionary}
Prahlad Vadakkepat, Kay~Chen Tan, and Wang Ming-Liang.
\newblock Evolutionary artificial potential fields and their application in
  real time robot path planning.
\newblock In {\em Proceedings of the 2000 congress on evolutionary computation.
  CEC00 (Cat. No. 00TH8512)}, volume~1, pages 256--263. IEEE, 2000.

\bibitem{verma2018feastnet}
Nitika Verma, Edmond Boyer, and Jakob Verbeek.
\newblock Feastnet: Feature-steered graph convolutions for 3d shape analysis.
\newblock In {\em Proceedings of the IEEE conference on computer vision and
  pattern recognition}, pages 2598--2606, 2018.

\bibitem{wang2018deep}
Shenlong Wang, Simon Suo, Wei-Chiu Ma, Andrei Pokrovsky, and Raquel Urtasun.
\newblock Deep parametric continuous convolutional neural networks.
\newblock In {\em Proceedings of the IEEE Conference on Computer Vision and
  Pattern Recognition}, pages 2589--2597, 2018.

\bibitem{wang2018sgpn}
Weiyue Wang, Ronald Yu, Qiangui Huang, and Ulrich Neumann.
\newblock Sgpn: Similarity group proposal network for 3d point cloud instance
  segmentation.
\newblock In {\em Proceedings of the IEEE Conference on Computer Vision and
  Pattern Recognition}, pages 2569--2578, 2018.

\bibitem{wang2019dynamic}
Yue Wang, Yongbin Sun, Ziwei Liu, Sanjay~E Sarma, Michael~M Bronstein, and
  Justin~M Solomon.
\newblock Dynamic graph cnn for learning on point clouds.
\newblock {\em Acm Transactions On Graphics (tog)}, 38(5):1--12, 2019.

\bibitem{wu2019pointconv}
Wenxuan Wu, Zhongang Qi, and Li Fuxin.
\newblock Pointconv: Deep convolutional networks on 3d point clouds.
\newblock In {\em Proceedings of the IEEE Conference on Computer Vision and
  Pattern Recognition}, pages 9621--9630, 2019.

\bibitem{wu20153d}
Zhirong Wu, Shuran Song, Aditya Khosla, Fisher Yu, Linguang Zhang, Xiaoou Tang,
  and Jianxiong Xiao.
\newblock 3d shapenets: A deep representation for volumetric shapes.
\newblock In {\em Proceedings of the IEEE conference on computer vision and
  pattern recognition}, pages 1912--1920, 2015.

\bibitem{xu2018spidercnn}
Yifan Xu, Tianqi Fan, Mingye Xu, Long Zeng, and Yu Qiao.
\newblock Spidercnn: Deep learning on point sets with parameterized
  convolutional filters.
\newblock In {\em Proceedings of the European Conference on Computer Vision
  (ECCV)}, pages 87--102, 2018.

\bibitem{ye20183d}
Xiaoqing Ye, Jiamao Li, Hexiao Huang, Liang Du, and Xiaolin Zhang.
\newblock 3d recurrent neural networks with context fusion for point cloud
  semantic segmentation.
\newblock In {\em Proceedings of the European Conference on Computer Vision
  (ECCV)}, pages 403--417, 2018.

\bibitem{yi2016scalable}
Li Yi, Vladimir~G Kim, Duygu Ceylan, I-Chao Shen, Mengyan Yan, Hao Su, Cewu Lu,
  Qixing Huang, Alla Sheffer, and Leonidas Guibas.
\newblock A scalable active framework for region annotation in 3d shape
  collections.
\newblock {\em ACM Transactions on Graphics (ToG)}, 35(6):1--12, 2016.

\bibitem{yi2017syncspeccnn}
Li Yi, Hao Su, Xingwen Guo, and Leonidas~J Guibas.
\newblock Syncspeccnn: Synchronized spectral cnn for 3d shape segmentation.
\newblock In {\em Proceedings of the IEEE Conference on Computer Vision and
  Pattern Recognition}, pages 2282--2290, 2017.

\bibitem{zaheer2017deep}
Manzil Zaheer, Satwik Kottur, Siamak Ravanbakhsh, Barnabas Poczos, Russ~R
  Salakhutdinov, and Alexander~J Smola.
\newblock Deep sets.
\newblock In {\em Advances in neural information processing systems}, pages
  3391--3401, 2017.

\bibitem{zhang2018efficient}
Chris Zhang, Wenjie Luo, and Raquel Urtasun.
\newblock Efficient convolutions for real-time semantic segmentation of 3d
  point clouds.
\newblock In {\em 2018 International Conference on 3D Vision (3DV)}, pages
  399--408. IEEE, 2018.

\bibitem{zhang2019shellnet}
Zhiyuan Zhang, Binh-Son Hua, and Sai-Kit Yeung.
\newblock Shellnet: Efficient point cloud convolutional neural networks using
  concentric shells statistics.
\newblock In {\em Proceedings of the IEEE/CVF International Conference on
  Computer Vision}, pages 1607--1616, 2019.

\end{thebibliography}
}

\end{document}